\documentclass{article}

\usepackage[numbers]{natbib} 
\usepackage[final]{neurips_2025}

\usepackage{algorithm}
\usepackage{algpseudocode}
\usepackage{amsmath}
\usepackage{graphicx} 
\usepackage{subcaption}

\usepackage[utf8]{inputenc} 
\usepackage[T1]{fontenc}    
\usepackage{hyperref}       
\usepackage{url}            
\usepackage{booktabs}       
\usepackage{amsfonts}       
\usepackage{nicefrac}       
\usepackage{microtype}      
\usepackage{xcolor}         
\usepackage{graphicx, wrapfig}

\title{Skrull: Towards Efficient Long Context Fine-tuning through Dynamic Data Scheduling}

\begin{document}

%

%

\author{
Hongtao Xu$^{1,2,3}$ \quad Wenting Shen$^{3}$ \quad Yuanxin Wei$^{4}$ \quad Ang Wang$^3$ \quad Guo Runfan$^{2}$ \\
\textbf{Tianxing Wang}$^{3}$ \quad \textbf{Yong Li}$^{3}$ \quad \textbf{Mingzhen Li}$^{2\dag}$ \quad \textbf{Weile Jia}$^{2\dag}$ \\
$^1$School of Advanced Interdisciplinary Sciences, 
University of Chinese Academy of Sciences \\
\quad $^2$State Key Lab of Processors, Institute of Computing Technology, CAS \\
\quad $^3$Alibaba Group \\
\quad $^4$Sun Yat-sen University \\
$^{\dag}$ Corresponding authors
}


\maketitle

\vspace{-5mm}
\begin{abstract}
Long-context supervised fine-tuning (Long-SFT) plays a vital role in enhancing the performance of large language models (LLMs) on long-context tasks. To smoothly adapt LLMs to long-context scenarios, this process typically entails training on mixed datasets containing both long and short sequences. However, this heterogeneous sequence length distribution poses significant challenges for existing training systems, as they fail to simultaneously achieve high training efficiency for both long and short sequences, resulting in sub-optimal end-to-end system performance in Long-SFT.
In this paper, we present a novel perspective on data scheduling to address the challenges posed by the heterogeneous data distributions in Long-SFT. We propose Skrull, a dynamic data scheduler specifically designed for efficient long-SFT. Through dynamic data scheduling, Skrull balances the computation requirements of long and short sequences, improving overall training efficiency. Furthermore, we formulate the scheduling process as a joint optimization problem and thoroughly analyze the trade-offs involved. Based on those analysis, Skrull employs a lightweight scheduling algorithm to achieve near-zero cost online scheduling in Long-SFT. Finally, we implement Skrull upon DeepSpeed, a state-of-the-art distributed training system for LLMs. Experimental results demonstrate that Skrull outperforms DeepSpeed by 3.76x on average (up to 7.54x) in real-world long-SFT scenarios.

\end{abstract}

\vspace{-5mm}
\section{Introduction}
\label{sec:intro} 
Long-context capabilities are important for large language models (LLMs) to handle various tasks such as long document summarization, question answering, multi-turn dialogue and code generation. Mainstream LLMs such as Llama \cite{llama2_tech_report_2023,llama3_tech_report_2024}, Qwen \cite{qwen25_tech_report_2024} and GPT-4 \cite{gpt4_tech_report_2024} can support the context window of up to 128K tokens. Google's Gemini \cite{gemini15_tech_report_2024} can even achieve up to 1M tokens per context window. Typically, additional training phases like long-context supervised fine-tuning (Long-SFT) as well as long-context continue pre-training (Long-CPT) are employed to extend the context length. For example, Llama3 \cite{llama3_tech_report_2024} is fine-tuned with 99.89\% short sequence (averaging under 1K tokens) and 0.11\% long sequence (averaging around 37K tokens). Qwen2.5-Turbo~\cite{qwen25_tech_report_2024} gradually extends context length by training on 40\% long sequences and 60\% short sequences. Training on those meticulous gathered datasets enables smoothly adaptation of LLMs to longer context while still maintaining the performance on short context tasks. 

However, this heterogeneous data distribution in Long-SFT poses significant challenges for existing distributed LLM training frameworks \cite{megatron_v3_sp_selective_recompute_2023, deepspeed_zero_sc20, pytorch_distributed}, exhibiting sub-optimal efficiency. For instance, the heterogeneous data distribution poses a dilemma for parallelism and memory-reduction strategies. Specifically, long sequences necessitate context parallelism and other memory-reduction approaches due to their tremendous memory requirements. However, those approaches compromise the training efficiency for short ones due to the overheads like unnecessary communication and GPU under-utilization. Moreover, the wide sequence length distribution in long-SFT worsen the mismatch of computation characteristics in \texttt{Attention} module, which exhibit quadratic computational complexity and linear memory consumption \cite{flashattention_v1_2022,flashattention_v2_2024}, leading to another dilemma for load balance problem.

To tackle the above challenges, we propose Skrull, a dynamic data scheduler dedicated for Long-SFT scenarios. Skrull efficiently handle the unique data distributions in Long-SFT scenario through two main components: Distributed-Aware Context Parallelism (DACP) and Global Data Scheduling (GDS). DACP selectively shards sequences and schedules them across different workers to minimize the performance degradation while maintains the ability of handling long sequence. GDS enlarge the scope of scheduling and improve the GPU utilization during training. The two components collaborate with each other at different scheduling granularities. Furthermore, to achieve the optimal performance, we formulate the scheduling process as a joint optimization problem and design a lightweight heuristic algorithm to solve it at runtime. Experimental results demonstrate that Skrull improves the end-to-end training performance by 3.76x on average (up to 7.54x) compared to DeepSpeed, a state-of-the-art distributed LLM training framework. 

Our key contributions are summarized as follows:
\begin{itemize}
    \item We provide a new perspective of data scheduling to address the heterogeneous sequence length distribution.
    \item We propose a new context parallelism called DACP based on fine-grained data scheduling, which maintaining both the processing capabilities for long sequences and efficiency for short sequences, enabling efficient training on heterogeneous data distribution in long-SFT scenario.
    \item We implement coarse-grained global data scheduling (GDS) and further formulate GDS and DACP as a joint optimization problem through performance modeling.
    \item We design a lightweight heuristic algorithm and achieve performance gains by 3.76x on average (with a peak improvement of 7.54×) in real-world datasets.
\end{itemize}


\section{Preliminaries}   
\paragraph{Data Parallelism (DP).} Data parallelism \cite{pytorch_distributed, pytorch_fsdp, deepspeed_zero_sc20} partitions the training samples to multiple workers and each worker maintains a complete model weight replica. In each iteration, all workers process a subset of global batch independently and then synchronize the gradients across all DP ranks. 
However, due to the inherent synchronization semantic in DP, the load balance becomes a noticeable problem, especially in long context scenarios.
 


\paragraph{Context Parallelism (CP).} \label{sec:cp} 
Context parallelism partitions the input tensor along the sequence length dimension and distributes it to multiple workers \cite{deepspeed_ulysess_2024, ring_attention_2024, llama3_tech_report_2024}. CP is emerging as an inevitable parallel strategy when handling long context. 
In the Transformer architecture, the primary challenge of CP stems from the parallelization of \texttt{Attention} module because each tokens needs to attend to other tokens in the sequence. Consequently, the communication in CP is inevitable. Notably, DACP, proposed in this paper, leverages data scheduling to minimize the overheads caused by CP and is orthogonal to specific CP implementations.

\section{Observation}
\label{sec:motivation} 
\subsection{Heterogeneous Sequence Length Distribution}
\label{sec:motivation_skewed_dataset}
\begin{figure}[htbp]
    \begin{subfigure}[t]{0.48\textwidth}
        \centering
        \includegraphics[width=\linewidth]{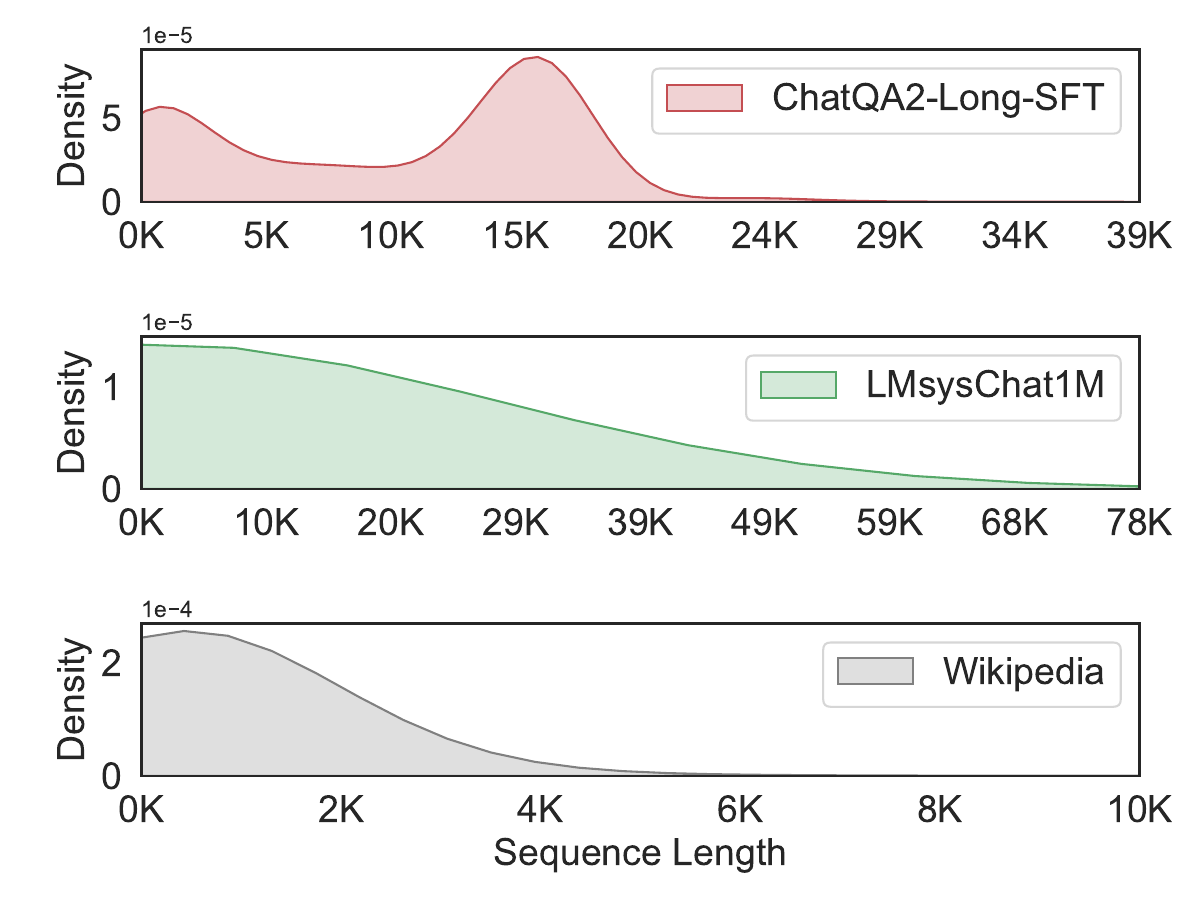}%
        \caption{Sequence length distribution.}
        \label{fig:data_distribution}
    \end{subfigure}%
    \hfill%
    \begin{subfigure}[t]{0.48\textwidth}
        \centering
        \includegraphics[width=\linewidth]{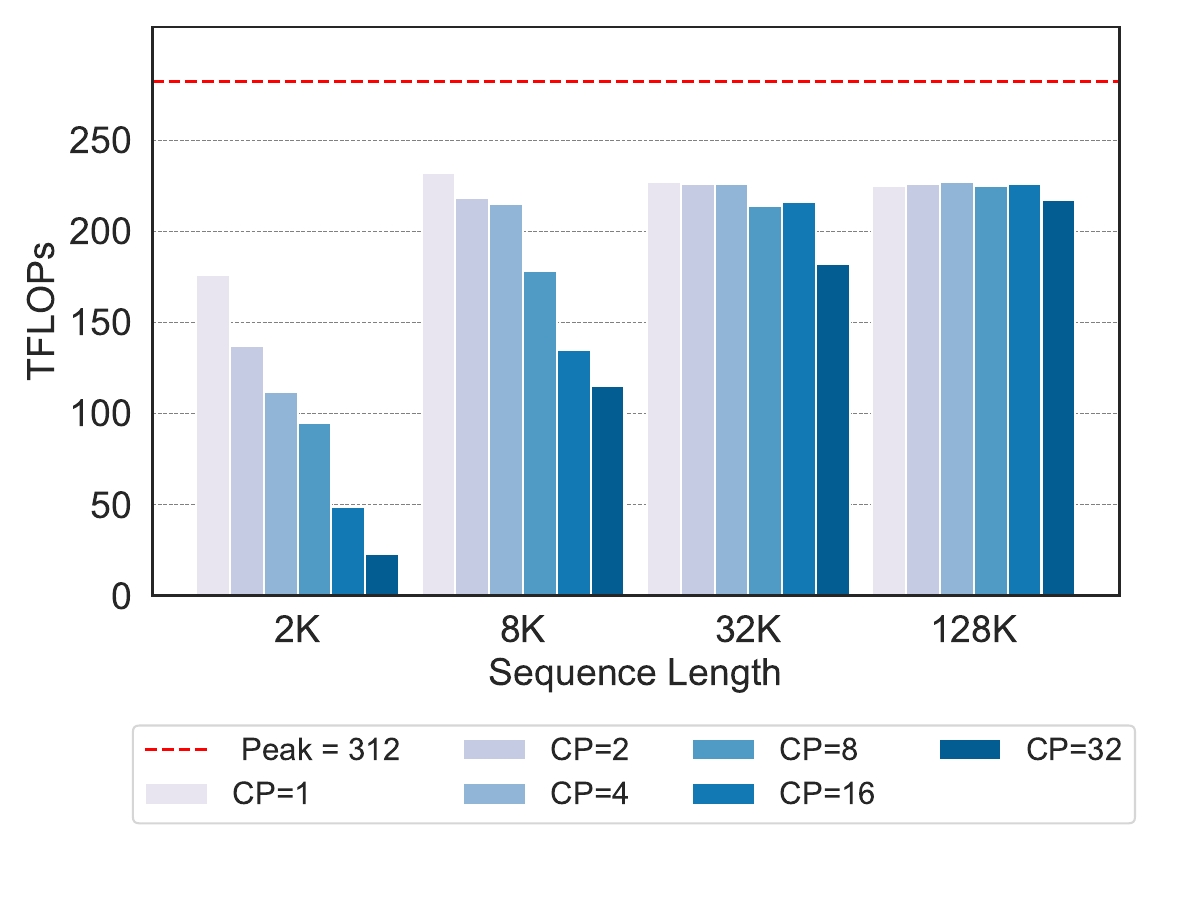}
        \caption{Performance (FLOPS) with different CP workers.}
        \label{fig:flops_cp}
    \end{subfigure}
    \caption{Sequence length distribution on different datasets, and corresponding performance impact.}
\end{figure}

As shown in Figure~\ref{fig:data_distribution}, we observe pronounced variance in the sequence length distribution across real-world Long-SFT datasets, including Wikipedia \cite{dataset_wikipedia}, LMsysChat1M \cite{dataset_lmsyschat1m} and ChatQA2-Long-SFT \cite{dataset_chatqa2-long-sft-data_2025}. Among them, the sequence length distribution of ChatQA2-long-SFT exhibits a bimodal pattern, where the proportions of long and short sequences are nearly equal. Specifically, approximately 40\% of sequences are shorter than 8K tokens, while the remaining 60\% exceed this threshold. As comparison, long-tail distributions represent another typical pattern in Long-SFT datasets. In Llama3's internally collected Long-SFT datasets \cite{llama3_tech_report_2024}, we find that 99.89\% of sequences are under 1K tokens on average, while the remaining 0.11\% are approximately 37K tokens, showcasing extremely skewed long-tail distribution. Due to data accessibility constraints, we plot the sequence length distribution of Wikipedia and LMsysChat1M in Figure~\ref{fig:data_distribution}, which have the identical feature with Llama3's Long-SFT dataset. Table~\ref{tab:token_length} lists the portions under different lengths thresholds for these three datasets, highlighting the heterogeneous sequence length distribution in Long-SFT.

\begin{table}[htbp]
\centering
\caption{Percentage of sequence length in real-world datasets.}
\label{tab:token_length}
\begin{tabular}{lcccccc}
\toprule
\textbf{Dataset} & \textbf{<1K} & \textbf{<4K} & \textbf{<8k} & \textbf{<32K} & \textbf{<128K} & \textbf{Longest} \\
\midrule
Wikipedia & 87.88\% & 99.34\% & 99.92\% & 99.99\% & 100.0\% & 78K \\
LMsysChat1M & 87.12\% & 99.35\% & 99.87\% & 99.98\% & 99.99\% & 1643K \\
ChatQA2-Long-SFT & 21.92\% & 31.48\% & 40.43\% & 99.86\% & 100.0\% & 99K \\
\bottomrule
\end{tabular}
\end{table}

\subsection{Performance Degradations for Short Sequences}
\label{sec:motivation_performance_degradations}
In this section, we discuss our observation on the performance degradations and GPU under-utilization for short sequences in Long-SFT training. During the training process, the context parallelism degree and other memory reduction strategies such as gradient accumulation are set to accommodate the longest sequence in datasets to avoid out-of-memory errors (OOMs). However, these training settings degrade their performance for the shorter sequences, which make up the majority in Long-SFT datasets. As shown in Figure~\ref{fig:flops_cp}, we test the performance of \texttt{Attention} module \cite{flashattention_v2_2024} under different CP degrees. Results demonstrate, especially for the short sequences, higher CP degree exacerbates kernel execution efficiency. Additionally, context parallelism also brings unnecessary communication overhead to short sequences. Also, the memory reduction strategies tailored to long sequence lead to low GPU memory utilization for the most time.

\section{Skrull} 
\begin{figure}[t]
    \centering
    \includegraphics[width=1.0\linewidth]{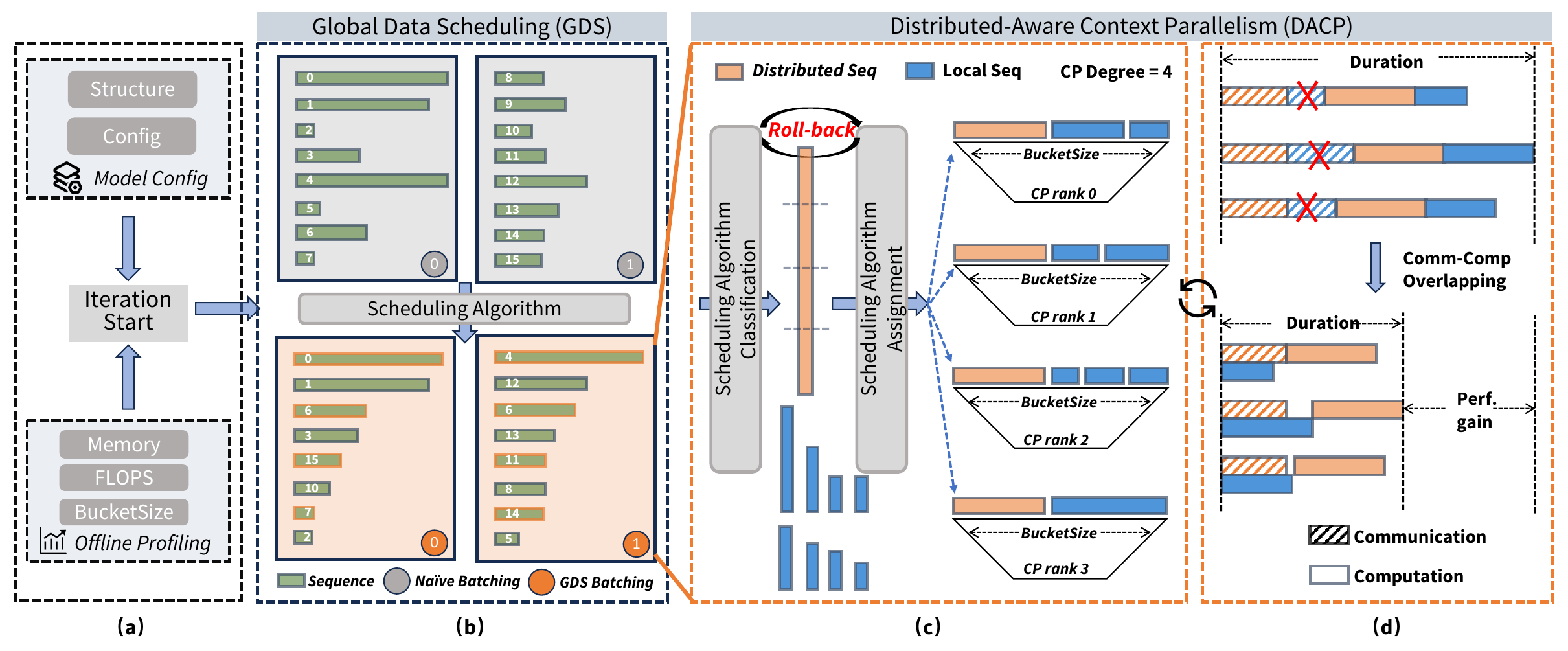} 
    \caption{Workflow of Skrull. (a) Offline profiling: Given model and training settings, it provides performance estimation for data scheduling. (b) GDS: produce optimal batching strategies for DACP. (c) DACP: dynamically scheduling data to specific hardware with balanced workload and minimum overheads. (d) Performance gains of DACP: it shows how the reduced communication volumn and overlapping improve the performance.} 
    \label{fig:Skrull_overview}
\end{figure}

We introduce design of Skrull and the efficient implementation for online Long-SFT training in this section. Figure~\ref{fig:Skrull_overview} illustrates the workflow of Skrull. From the perspective of data scheduling, Skrull consists of two parts: (i) Global data scheduling (GDS): For every iteration, Skrull takes the global batch as input and employs coarse-grained scheduling to generate the optimal micro-batches for each DP ranks. (ii) Distributed-aware Context Parallelism (DACP): Taking the micro-batch produced in GDS, Skrull further employs finer-grained scheduling to selectively distribute the sequences and assign them to different CP workers. For the convenience of formulation, we sequentially introduce DACP in Section~\ref{sec:DACP}, GDS in  Section~\ref{sec:GDS} and the efficient implementations in Section~\ref{sec:efficient_solver}.

\subsection{Distributed-aware Context Parallelism}
\label{sec:DACP}
To simultaneously achieve high efficiency for all the sequences, we propose \textbf{d}istributed-\textbf{a}ware \textbf{c}ontext \textbf{p}arallelism (DACP). As shown in Figure~\ref{fig:Skrull_overview}(c), DACP dynamically determines whether to distribute the sequences to avoid unnecessary overheads or not. 
On the one hand, DACP preserves the original context parallel settings to maintain the ability of handling long sequences. 
On the other hand, DACP selectively schedules short sequences entirely within a single device to minimize the degradation. 
Therefore, based on distinct computational characteristics, DACP classifies sequences into two categories: (i) distributed sequences requiring context parallelism, and (ii) local sequences needing efficient processing and intended to reside entirely within a single device.
Notably, these sequences are still processed within a shared CP group without increasing the number of GPUs used for training. 
Furthermore, as illustrated in Figure~\ref{fig:Skrull_overview}(d), DACP brings an additional opportunity to overlap the communication of distributed sequences and the computation of local sequences in \texttt{Attention} module due to the inherent independence between distributed and local sequences. 

However, the scheduling process presents significant challenges. First, inappropriate sequence classification may lead to out-of-memory errors (OOMs). Second, the local sequences are varying in length and pose load imbalance issue across CP ranks. To fully explore the relationship between scheduling plans and performance gains, we first analyze the computation and memory features in Appendix~\ref{sec:apdx_performance_modeling}. Through offline profiling, we model the computation (see in Appendix~\ref{sec:apdx_compuation_estimator}) by $FLOPs$ function and latency estimation function $T_{comp}$. Additionally, we map the sequence length to the memory consumption and derive the BucketSize $C$ which indicates the capacity of total sequence length per ranks. The BucketSize $C$ plays a vital role in measuring the memory constrain during Skrull's scheduling. More details are listed in Appendix~\ref{sec:apdx_bucketsize}. Similarly, we model communication volume function $Volume$ and latency function $T_{comm}$, as detailed in Appendix~\ref{sec:apdx_comm_estimator}. Finally, we formulate the scheduling process as an optimization problem as follows. The frequently used notions are listed in Table~\ref{tab:notation}.  

\begin{table}[t]
\caption{Symbols used in this paper.}
\label{tab:notation}
\vskip 0.15in
\begin{center}
\begin{small}
\begin{tabular}{l p{5cm} l p{5cm}}
\toprule
\textbf{Symbol} & \textbf{Description} & \textbf{Symbol} & \textbf{Description} \\
\midrule
$S_k$   & Length of the $k$-th sequence in a batch. & $C$ & BucketSize per rank. \\
$D_k$    & Distribute $k$-th sequence or not. & $P_{kj}$ & Assign $k$-th sequence to CP rank $j$ \\
$N$ & CP degrees. & $FLOPs$ & FLOPs estimation function. \\
$T_{comp}$ & Computation cost estimation. & $T_{comm}$ & Communication cost estimation. \\
$B_kij$ & Assign $k$-th sequence to $i-th$ DP rank and $j-th$ micro-batch. & $ws$ & DP degree \\
$Volume$ & Communication volume count function. & & \\
\bottomrule
\end{tabular}
\end{small}
\end{center}
\vskip -0.1in
\end{table}

\paragraph{DACP Formulation.}
\label{sec:DACP_formulations}
We first define the sequence classification array $D \in \{0,1\}^K $ ( 0 for local sequence and 1 for distributed sequence) and local sequence assignment matrix $P \in \{0,1\}^{K \times N}$ ( 1 for assignment and 0 for not). For example, $D_k = 1$ indicates that the $k-th$ sequence with length of $S_k$ is scheduled to be computed in a distributed manner. Similarly, $P_{kj} = 1$ indicates that the $k-th$ sequence is assigned to device $j$, implying $D_k = 0$. Given a micro-batch comprising $K$ sequences with lengths $S_{k}$ ($k = (0, \ldots, K-1)$), BucketSize $C$ and CP degree $N$, the scheduling process of DACP can be formulated as follows:
{\footnotesize 
\begin{alignat}{2}
    & \text{\textbf{Objective}} \quad && \min_{arg} \textbf{TDACP} = \min_{arg} \max_{j} \left( \text{Time}_j \right) \label{eq:DACP_obj} \\
    & \text{\textbf{Subject to}} \quad && \nonumber \\
    & && \text{Time}_j = \max \left( T_{comm}\left(\text{V}\right), T_{comp} \left(  \text{Local}_j\right) \right) + T_{comp} \left( \text{Dist} \right), \quad \forall j \label{eq:DACP_cost_time} \\ 
    & && \text{Local}_j = \sum_k \text{FLOPs} \left( P_{kj} \cdot S_k \right), \quad \forall j \label{eq:DACP_local_flops} \\
    & && \text{Dist} = \frac{1}{\text{N}} \sum_k \text{FLOPs} \left( D_k \cdot S_k \right) \label{eq:DACP_dist_flops} \\
    & && \text{V} = \text{Volume}( \sum_k D_k \cdot S_k) \label{eq:DACP_dist_volume} \\
    & && \sum_j P_{kj} + D_k = 1, \quad \forall k \label{eq:DACP_assign_complete} \\
    & && \sum_k S_k \cdot P_{kj} + \frac{D_k \cdot S_k}{\text{N}} \leq C, \quad \forall j \label{eq:DACP_assign_mem_sound}
\end{alignat}
}%
Here, our optimization goal is to find the optimal $D$ and $P$ to minimize $TDACP$, which represents the total duration in one micro-batch. As shown in Equation~\ref{eq:DACP_obj}, $TDACP$ is determined by the maximum execution time $\text{Time}_j$ across all CP ranks $j$. 
Specifically, as described in Equation~\ref{eq:DACP_cost_time}, $\text{Time}_j$ consists of two components: (1) the overlapping term, defined as the maximum of the communication time $T_{comm}(\text{V})$ and the computation time $T_{comp}(\text{Local}_j)$ for local sequences, and (2) the computation time $T_{comp}(\text{Dist})$ for distributed sequences. Here, $T_{comm}$ depends on the communication volume $\text{V}$, as modeled in Equation~\ref{eq:DACP_dist_volume}. Similarly, $T_{comp}$ utilizes the results from Equations~\ref{eq:DACP_local_flops} and~\ref{eq:DACP_dist_flops}, which compute the FLOPs for local sequences on CP rank $j$ and distributed sequences, respectively.
Finally, Equation~\ref{eq:DACP_assign_complete} ensures the completeness of data scheduling, while Equation~\ref{eq:DACP_assign_mem_sound} enforces the memory constraint.

\vspace{-0.1cm}
\subsection{Global Data Scheduling} 
\label{sec:GDS}
Section~\ref{sec:DACP} discusses the data scheduling in the scope of micro-batch. However, only relying on scheduling in DACP is insufficient. The reasons are as follows.

First, the heterogeneous sequence length distribution also leads the to severe load imbalance across different micro-batches, resulting in the sub-optimal training efficiency in Long-SFT scenarios. Second, to achieve maximum performance gains in DACP, meticulous micro-batching strategy is essential. For example, pairing long and short sequences with appropriate memory pressure can expand the valid scheduling space for DACP. Specifically, micro-batches with large total sequence lengths increase the risk of OOMs and limit the optimizations in DACP, such as selective sharding. In contrast, micro-batches with small total sequence lengths introduce GPU under-utilization, degrading the end-to-end performance. Therefore, as shown in Figure~\ref{fig:Skrull_overview}(b), Skrull employs Global Data Scheduling (GDS), which derives the optimal micro-batching strategy from the global batch. We limit the scheduling scope to the global batch because it represents the maximum scope that maintains mathematical equivalence for mainstream optimizers such as Adam \cite{optimizer_adam_2014} and AdamW \cite{optimizer_adamw_2017}. 

\paragraph{Joint Formulation}
\label{sec:joint_formulations}
We re-formulate the scheduling process as a joint optimization problem that integrates both DACP and GDS. We first define the batching matrix $B_{kij} \in \{0,1\}^{K \times N}$, which indicates whether the $k$-th sequence is scheduled into the $j$-th micro-batch of DP rank $i$.
Given a global batch $B$ consisting of $K$ sequences with lengths $S_k$, we re-formulate the scheduling process as follows:
{\footnotesize 
\begin{alignat}{2}
    & \text{\textbf{Objective}} \quad && \min_{arg} \max_{i} ( \sum_j Time_{ij}) \label{eq:opt_gds_obj} \\
    & \text{\textbf{Subject to}} \quad && \nonumber \\
    & && \sum_{ij} B_{kij}= 1, \quad \forall k \label{eq:opt_gds_norm} \\
    & && \sum_{ki} B_{kij} * S_k \leq C * N, \quad \forall j \label{eq:opt_gds_mem} \\
    & && \text{Time}_{ij} = \textbf{TDACP} \left( B_{kij} * S_k\right), \quad \forall k \label{eq:opt_gds_time} 
\end{alignat}
}%

Here, \textbf{TDACP} represents the function in Equation~\ref{eq:DACP_obj}. Equation~\ref{eq:opt_gds_norm} ensures all sequences are assigned exactly once. The memory constraint in Equation~\ref{eq:opt_gds_mem} prevents the OOMs while the $\text{Time}_{ij}$ shown in Equation~\ref{eq:opt_gds_time} provides cost estimations for each micro-batch using DACP formulation in Equation~\ref{eq:DACP_obj}. As shown in Equation~\ref{eq:opt_gds_obj}, the total execution duration per iteration is determined by the DP rank with the longest cumulative execution time across its micro-batches.

Overall, the optimization target is to the minimize the total execution time per iteration by deducing the optimal scheduling plan, which is represented by a combination of $B_{kij}$, $D_k$ and $P_{kj}$.

\subsection{Efficient Online Scheduling}
\label{sec:efficient_solver} 
Although some solvers like \cite{SCIP_MILP_Sovler} can derive the optimal scheduling plan, its long solving time makes it impractical for scheduling during runtime. To achieve online scheduling during Long-SFT, we resort to design lightweight heuristic scheduling algorithm. Notably, our scheduling algorithm is integrated into the DataLoader and introduces near-zero overhead to the training process.

\subsubsection{Memory vs. Computation: Trade-off Analysis}
\label{sec:trade-off_analysis}
\textbf{Memory} and \textbf{Computation} are the key factors related to the performance, as shown in the formulations of scheduling.
We should achieve optimal performance while not violating memory constraint, presenting a trade-off. 
Therefore, we first analyze the trade-off between Computation and Memory when deducing the scheduling strategies, highlighting the considerations when designing the scheduling algorithms. 
\paragraph{Sequence classification: deduce the array $D$.}
We analyze the sequence classification (array $D$ in Section~\ref{sec:DACP_formulations}). 
From the perspective of \textbf{Computation}, $D$ impacts the communication volume and computation of sharded sequences (Equation~\ref{eq:DACP_dist_volume} and~\ref{eq:DACP_dist_flops}). 
More sharded sequences will incur more performance degradation, which comes from both communication overhead and kernel execution (refer to Section~\ref{sec:motivation_performance_degradations}).
However, from the perspective of \textbf{Memory}, more distributed sequences will bring more balanced memory consumption (Equation~\ref{eq:DACP_assign_mem_sound}), which can lower the risk of OOMs, as the remaining local sequences with varying lengths are hard to be assigned evenly.
Besides, although the overlapping in DACP can alleviate the performance degradation problem to some extent (Equation~\ref{eq:DACP_cost_time}), it is still non-trivial to decide the optimal classification array $D$.

\paragraph{Local sequence assignment: deduce the matrix $P$.}

Then, we analyze local sequence assignment, which is represented by $P$.
From the perspective of \textbf{Computation}, $P$ impacts the Equation~\ref{eq:DACP_local_flops}, which implies the computation workload in each CP ranks, thus affects the load balance. The ideal situation is to balance the local sequences for computation balance among CP ranks.However, from the perspective of \textbf{Memory}, the scheduling which balances the computation leads to the unbalance of memory consumption, which increases the risk of OOMs.

Unfortunately, we cannot balance the computation and memory at the same time. 
The reason is that, after applying FlashAttention~\cite{flashattention_v1_2022,flashattention_v2_2024}, the correlation between computation complexity and and sequence length ($n$) is $O(n^2)$, however, the correlation between memory is $O(n)$. 
Moreover, with the sequence length increasing, the portion of \texttt{Attention} module gradually dominates the computation load, making it more difficult to balance the computation and memory. 
Worse still, the model configuration (e.g., KV heads, hidden size) also impacts. Due to the limited page, we list the details in Appendix~\ref{sec:apdx_performance_modeling}.



Therefore, we need to carefully deal with the memory footprint balance and the computation complexity balance and we design the following heuristics. 

\subsubsection{Heuristics}
\label{sec:heuristics}
\paragraph{Scheduling Algorithm of DACP.}
We first summarize three principles of algorithm design in DACP. (i) \textbf{Avoid sharding}: We strive to avoid sequence sharding and assume that all sequences will be handled locally first. (ii) \textbf{Prioritize computation}: We prioritize balancing computation over memory to achieve better performance. (iii) \textbf{Roll-back mechanism}: We continuously monitor the estimated memory consumption and revert decisions when necessary.   
The roll-back mechanism guarantees the memory constrains outlined in Equation~\ref{eq:DACP_assign_mem_sound} and Equation~\ref{eq:opt_gds_mem}, while enabling more aggressive scheduling attempts based on (i) and (ii). 
Our heuristic for DACP is listed in Algorithm~\ref{alg:heuristic_cp}. Given a micro-batch containing $K$ sequences with lengths $S[K]$ and a predefined BucketSize $C$, the algorithm outputs the sequence classification and assignment results in the form of an array $ret$. In this array, a value of -1 at the i-th position indicates that the i-th sequence is to be sharded, while a value $v = (0, \ldots, ws-1)$ indicates that the i-th sequence is assigned to CP rank $v$ entirely.
To better balance computation while ensuring memory constrains, we maintain two arrays during DACP scheduling: RemainBucket $RB$ and Loads $L$, which represent the current memory budget and computation load, respectively.  
We first sort the sequences in ascending order. For the each sequence, we sequentially assign it to the bucket (as well as CP rank) with minimum $L$ to avoid sharding and prioritize balancing computation (line 6-8). If the bucket cannot accommodate the sequence, we attempt to assign it to the bucket with the maximum $RB$ to avoid sharding (line 10-12). If both attempts fail, we classify the sequence as a distributed sequence and attempt to shard it (line 14-16). However, if the bucket with minimum of $RB$ cannot handle the sub-sequence after sharding, this indicates that the earlier process incorrectly classified inappropriate sequences as local sequences within this bucket. To address this, we employ a roll-back mechanism (line 18 and Appendix~\ref{apdx:details_on_heuristic_CP}). This mechanism identifies a local sequence in the bucket, shards it to reduce memory pressure, and resumes the assignment process. If the roll-back fails due to the absence of local sequences in the bucket, we return a DACP scheduling error. In such cases, GDS will also revert the batching plan (see the Section~\ref{sec:heuristic_GDS}). Notably, every assignment updates $RB$ and $L$ through the predefined functions \textit{UpdateLocal} and \textit{UpdateAll}. The details of these functions including \textit{RollBack} are further elaborated in Appendix~\ref{apdx:details_on_heuristic_CP}.

\begin{algorithm}[h]
\footnotesize
\caption{Heuristic scheduling algorithm of DACP}
\label{alg:heuristic_cp}
\begin{algorithmic}[1]
\Require SeqNum $K$, SeqLens $S[K]$, BucketSize $C$, CP degree $N$
\Ensure Scheduling Result $ret[K]$ 
\State \textbf{Sort(SeqLens, ascending=True)}
\For{$i=0$ to $N-1$} 
    \State $RB[i]{\gets}C, L[i]{\gets}0$ \Comment{Initialization}
\EndFor
\For{$i=0$ to $K-1$}
   \State $t \gets \text{argmin}(L)$ \Comment{Find rank t with minimum workload}
   \If{$RB[t]{\geq}S[i]$}
      \State $ret[i]{\gets}t$, UpdateLocal$(i,t)$
   \Else 
      \State $t \gets \text{argmax}(RB)$ 
      \If{$RB[t]{\geq}S[i]$} 
         \State $ret[i]{\gets}t$, UpdateLocal$(i,t)$
      \Else
          \State $t \gets \text{argmin}(RB)$
          \If{$RB[t]{\geq}S[i]/N$}
             \State $ret[i]{\gets}{-}1$, UpdateAll$(i)$ \Comment{Distribute the sequence}
          \Else
             \State \textbf{Assert} RollBack$(t,RB,L)$
             \State $i{\gets}i{-}1$ \Comment{Roll-back to avoid OOMs}
             \State \textbf{continue}
          \EndIf
      \EndIf
   \EndIf
\EndFor
\State \Return $ret$
\end{algorithmic}
\end{algorithm}

\paragraph{Scheduling Algorithm of GDS.} 
\label{sec:heuristic_GDS}
Algorithm~\ref{alg:heuristic_dp} demonstrates the heuristic scheduling algorithm of GDS. Given a global batch containing $K$ sequences with lengths $S[K]$, DP world size $ws$ and DP rank $dp\_rank$, the algorithm returns the scheduling result $mbs$, which consists of multiple micro-batches as inputs for Algorithm~\ref{alg:heuristic_cp}. We summarize three principles in our algorithm design. (i) \textbf{Prioritize computation}: We prioritize balancing computation across DP workers. To achieve this, we estimate the FLOPs (Appendix~\ref{sec:apdx_compuation_estimator}) and employ a bin-packing algorithm to balance computational workloads at a coarse granularity (line 1). (ii) \textbf{Pair long and short sequences}: We sort the sequences within each DP rank and batch them in an interleaved manner (line 7). This approach ensures that long sequences are assigned more evenly across micro-batches. Additionally, each micro-batch contains several short sequences, enhancing both task overlapping and load balancing. (iii) \textbf{Improve memory utilization}: We estimate the total memory requirements and try to improve the concurrency with less number of micro-batches. Thanks to the roll-back mechanism (line 8), this method maximizes memory utilization while not increase the risk of OOMs. As shown in line 5, we gradually increase the number of micro-batches if the scheduling fails and requires a roll-back.
 
\begin{algorithm}[htbp!]
\small
\caption{Heuristic Scheduling Algorithm of GDS}
\label{alg:heuristic_dp}
\begin{algorithmic}[1]
\Require SeqNum $K$, SeqLens $S[K]$, BucketSize $C$, CP degree $N$, DP WorldSize $ws$, DP\_Rank $dp\_rank$
\Ensure Micro-batches $mbs$
\State $Bin[ws] \gets \text{Binpack}(ws, \text{FLOPs}(S[K]))$ \Comment{Coarse-fined balance}
\State $Subset \gets Bin[dp\_rank]$, $init \gets \lceil \text{Sum}(Subset)/C \times N \rceil - 1$ 
\State \textbf{Sort}($Subset$, ascending=True)
\While{$init \leq K+1$}
    \State $init \gets init + 1$, $mbs \gets []$
    \For{$j \gets 0$ to $init$}
        \State $mbs.\text{append}(Subset[j::init])$ \Comment{Pair long and short sequences}
        \If{$\text{Sum}(mbs[-1]) \geq C \times N \textbf{ or not } \text{scheduling\_in\_DACP}(mbs[-1])$}
            \State \textbf{Continue} \Comment{Rollback if overload or DACP sheduling fails}
        \EndIf
    \EndFor
\EndWhile
\State \Return $mbs$
\end{algorithmic}
\end{algorithm}

\section{Evaluation}
\paragraph{Experimental Setup.} 
We conduct experiments using a testbed consisting of 4 nodes interconnected via a high-performance InfiniBand network, with each node equipped with 8 Nvidia H100 GPUs connected via 900GB/s NVLink. Then, We implement Skrull on top of DeepSpeed, a state-of-the-art distributed LLM training system and enable Zero-2 optimization as our baseline. Additionally, we implement sorted batching method in LongAlign \cite{bai_longalign_2024} for more comparison, which sort the dataset by sequence length and select random consecutive groups for each batch to improve the long-SFT training efficiency. We evaluation our optimizations on Qwen2.5-0.5B and Qwen2.5-7B using the three real-world datasets described in Section~\ref{sec:motivation_skewed_dataset}. Although Wikipedia and LMsysChat1M are not specifically gathered for Long-SFT, we still choose them as our evaluation datasets due to their long-tail distribution, which is exactly identical to Meta's in-house Long-SFT dataset \cite{llama3_tech_report_2024}. 
In contrast, ChatQA2-long-SFT dataset \cite{dataset_chatqa2-long-sft-data_2025} is specifically gathered for Long-SFT and exhibits bimodal distribution of data length, which is also similar to the dataset mentioned in \cite{qwen25_tech_report_2024}.
Through offline profiling, we configure the BucketSize to 26K and 13K for Qwen2.5-0.5B and Qwen2.5-7B, respectively. Further details regarding BucketSize configuration can be found in Appendix~\ref{sec:apdx_bucketsize}.
All the experiments share the same training settings with <DP=4, CP=8, BatchSize=64>, zero-2 enabled and selective recomputation strategy except for training Qwen-2.5-7B with ChatQA2-long-SFT dataset. Due to the increased memory requirements, we adjust its parallel settings with <DP=2, CP=16, BatchSize=40>. The global batch size is equal to DP size multiplied by BatchSize. Due to the limited page, we list precision validation in Appendix~\ref{apdx:loss_curves}.

\begin{figure}
    \centering
    \includegraphics[width=1\linewidth]{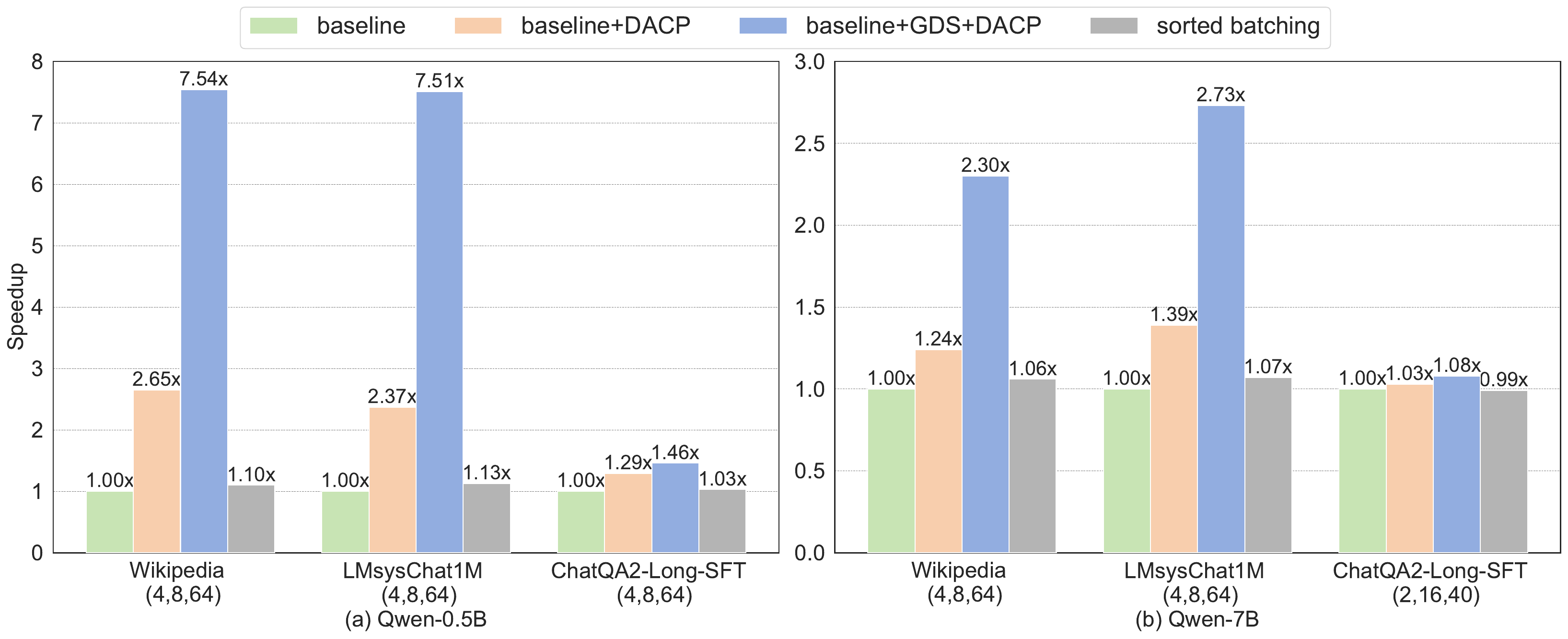}
    \caption{Overall performance and step-by-step evaluation. The settings represent the DP degree, CP degree and batch size, respectively.}
    \label{fig:experiment}
\end{figure}

\paragraph{Overall Performance.}
Figure~\ref{fig:experiment} illustrates the speedup achieved by Skrull, with the performance measured in terms of average iteration time. 
Skrull significantly outperforms the sorted batching strategy
Experimental results demonstrate that Skrull outperforms DeepSpeed and sorted batching method by an average of 3.76x and achieve peak improvement of 7.54×. The average speedups for Qwen-0.5B and Qwen-7B are 5.50x and 2.03x, respectively. We attribute this difference to the variation in BucketSize, which directly influences the valid data scheduling space. Also, Skrull outperforms sorted batching method by an averge of 3.45x with the peak improvement of 6.85x.
Additionally, from the perspective of datasets, the performance on Wikipedia and LMsysChat1M are similar due to the similar data distribution, which both exhibit long-tail feature. In this distribution, the short sequences dominate the datasets thus showcasing more optimization potential. In contrast, the long sequences also account for the majority in ChatQA2-Long-SFT dataset, which exhibits bimodal distribution, leading to relatively small optimization space. Specifically, when training Qwen-7B with this datasets, the major sequence length exceeds the BucketSize thus leading to limited speedup. We can further extend the BucketSize by combining more optimization techniques like parameter-efficient fine-tuning (PEFT) \cite{hu_lora_2022, chen_longlora_2024}. 

\paragraph{Step-by-step Evaluation.}
Additionally, we conduct step-by-step evaluation with the same training settings mentioned above. As shown in Figure~\ref{fig:experiment}, we successively enable DACP and GDS to test the effectiveness of each part in Skrull. Experimental results show that both components are effective and can cooperate well to further improve the end-to-end system performance in Long-SFT.

\paragraph{Performance Impact of BatchSize and BucketSize.} 

\begin{figure}
    \centering
    \includegraphics[width=1\linewidth]{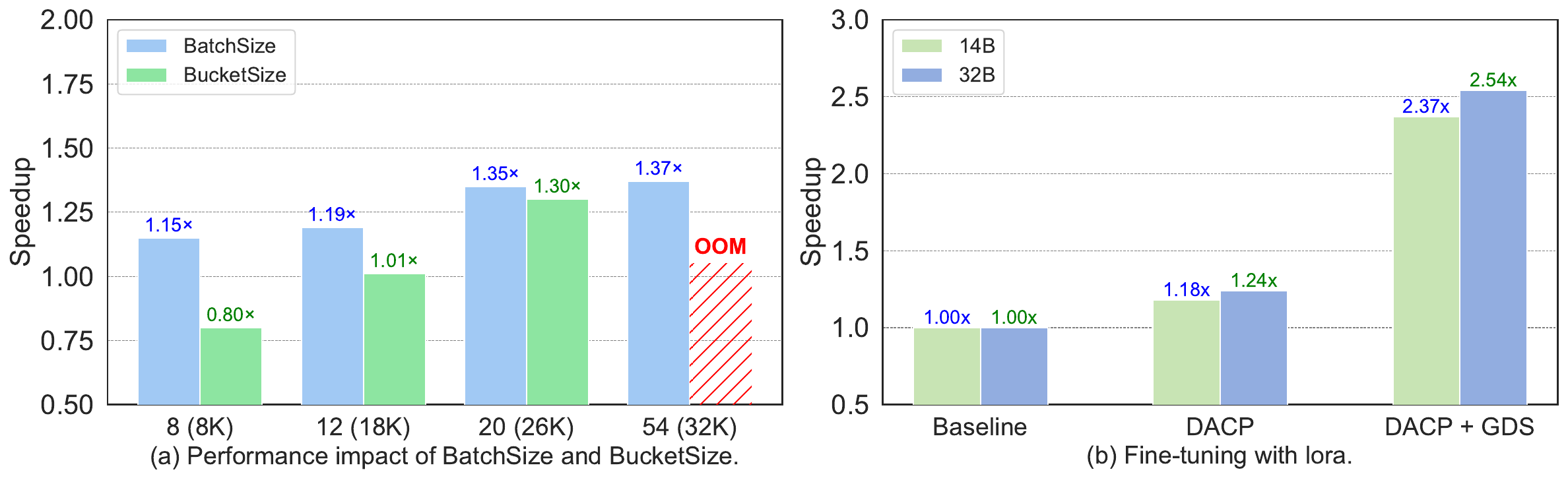}
    \caption{Left (a) shows the performance impact under different BatchSize and BucketSize. Right (b) shows Skrull's effectiveness and compatibility with Lora in larger models.}
    \label{fig:combined_performace_impact}
\end{figure}
To investigate the performance impact of BatchSize and BucketSize, we conduct experiments on ChatQA2-long-SFT using Qwen2.5-0.5B with the default setting of <DP=1, CP=8, BatchSize=64, BucketSize=26K>. As shown in Figure~\ref{fig:combined_performace_impact}(a), we adjust the BatchSize from 8 to 54 and the end-to-end speedup also improves. We attribute this performance gain to the expanded scheduling scope afforded by larger batch sizes. However, as the BatchSize increases further, the sampled batches gradually converge to the sequence length distribution of the dataset, causing the performance gains stabilized within a reasonable range. 
Additionally, we also evaluate the effect of BucketSize. Figure~\ref{fig:combined_performace_impact}(a) shows that increasing the BucketSize from 8K to 32K (values in parentheses) improves the speedup until an out-of-memory (OOM) error occurs. This indicates that while a larger BucketSize enhances performance, it also raises the risk of OOM errors. Therefore, it is important to set appropriate BucketSize, highlighting the importance of performance modeling module in Skrull.

\paragraph{Case Study}

\begin{table}[htbp]
  \centering
  \begin{minipage}[t]{0.48\textwidth}
    \centering
    \caption{Scheduling Strategies Comparison}
    \label{tab:scheduling_comparison}
    \begin{tabular}{@{} l c @{}}
      \toprule
      \textbf{Method} & \textbf{Speedup} \\
      \midrule
      RR w/ roll-back & 1.17$\times$ \\
      RR w/o roll-back & OOM \\
      Skrull w/ roll-back & 1.40$\times$ \\
      Skrull w/o roll-back & OOM \\
      \bottomrule
    \end{tabular}
  \end{minipage}%
  \hfill
  \begin{minipage}[t]{0.48\textwidth}
    \centering
    \caption{Latency Comparison per Iteration}
    \label{tab:memory_data}
    \begin{tabular}{@{} c c c c @{}}
      \toprule
      \textbf{Iteration} & \textbf{Baseline} & \textbf{RR} & \textbf{Skrull} \\
      \midrule
      1 & (36, 36) & (10, 49) & (21, 51) \\
      2 & (40, 40) & (29, 47) & (37, 50) \\
      3 & (35, 35) & (14, 45) & (32, 49) \\
      4 & (46, 46) & (46, 47) & (45, 49) \\
      \bottomrule
    \end{tabular}
  \end{minipage}
\end{table}

In this section, we present a quantitative analysis of the training process of Skrull. We conduct experiments using the Qwen2.5-0.5B model with the ChatQA2-Long-SFT dataset under the configuration <DP=1, CP=8, BatchSize=64>. To evaluate the effectiveness of our heuristics implementation, we compare it against a round-robin (RR) scheduling strategy, which assigns sequences in a simple round-robin manner (details in Appendix~\ref{apdx:RR_scheduling}). Additionally, we test both scheduling algorithms with and without roll-back mechanism to further assess the trade-off design in Skrull. 
As shown in Table~\ref{tab:scheduling_comparison}, Skrull significantly outperforms the RR scheduling. We analyze this result by examining quantitative data in the first four iterations. Table~\ref{tab:memory_data} (presented in tuple format) reports the minimum and maximum peak memory usage (in gigabytes) across all GPUs during each iteration. Compared to the baseline, both Skrull and RR scheduling allocate more sequences locally at the cost of increased memory imbalance. While such imbalance is acceptable as long as it dose not exceed memory capacity, it raises the risk of OOMs. 
Therefore, as shown in Table~\ref{tab:scheduling_comparison}, without the roll-back mechanism, both scheduling strategies result in OOMs, underscoring the importance of this safeguard.
In contrast to RR scheduling, Skrull achieves a better computational balance (indicated by the speedup) while adhering to memory constraints, demonstrating the effectiveness of its trade-off design in Skrull.

\section{Related Works} 
From the perspective of data engineering, those works \cite{zhao_longskywork_2024, bai_longalign_2024, qwen25_tech_report_2024, llama3_tech_report_2024} involve meticulously gathering training datasets for long context fine-tuning. From the perspective of training system, LongAlign \cite{bai_longalign_2024} adopts a sorted batching strategy to optimize system efficiency in long context fine-tuning phase. Chunkflow \cite{chunkflow} organize the training data into fixed size chunks, enabling controllable peak memory consumption and reduced pipeline bubbles. Additionally, some works employ dynamic parallelism settings \cite{hotspa_sosp24} to handle varying length sequences, which is similar to long-SFT. In contrast, Skrull adopts fixed parallelism settings and is orthogonal to those methods. Another type of works are parameter efficient finetuning (PEFT) \cite{hu_lora_2022, chen_longlora_2024} and Skrull is also effective for this methods.

\section{Conclusion}
In this paper, we provide a new perspective of data scheduling to enhance the training efficiency in Long-SFT scenarios. 
The heterogeneous data distribution in Long-SFT poses dilemmas for existing training systems on configuring parallelism strategies and ensuring the load balance. To tackle those challenges, we propose Skrull, a dynamic data scheduler dedicated for Long-SFT. Through dynamic data scheduling, Skrull achieves efficient training on both long sequences and short sequences. Additionally, we formulate the scheduling process as a joint optimization and adopt a lightwight scheduling algorithm. Experimental results demonstrate that Skrull outperforms DeepSpeed by 3.76x on average (up to 7.54x) in real-world long-SFT. Furthermore, we believe that Skrull can serve as an effective solution in other scenarios especially when dealing with mixture of long and short training data, such as reinforcement learning from human feedback (RLHF).

\section*{Acknowledgments}
This work is supported by the following funding: Beijing
Natural Science Foundation (4254087), National Science Foundation of
China (62502501, 92270206, 62372435), China National Postdoctoral
Program for Innovative Talents (BX20240383), Strategic Priority Research Program of Chinese Academy
of Sciences (XDB0500102), and Alibaba Research Intern Program. The model training were performed on the robotic AI-Scientist platform of Chinese Academy of Science and Alibaba Cloud Platform for AI (PAI).

\bibliography{nips_cite,framework,mainstream_llm} 


\medskip


\newpage
\appendix

\section{Precision Validation}
\label{apdx:loss_curves}

\begin{figure}[H]
    \centering
    \includegraphics[width=0.5\linewidth]{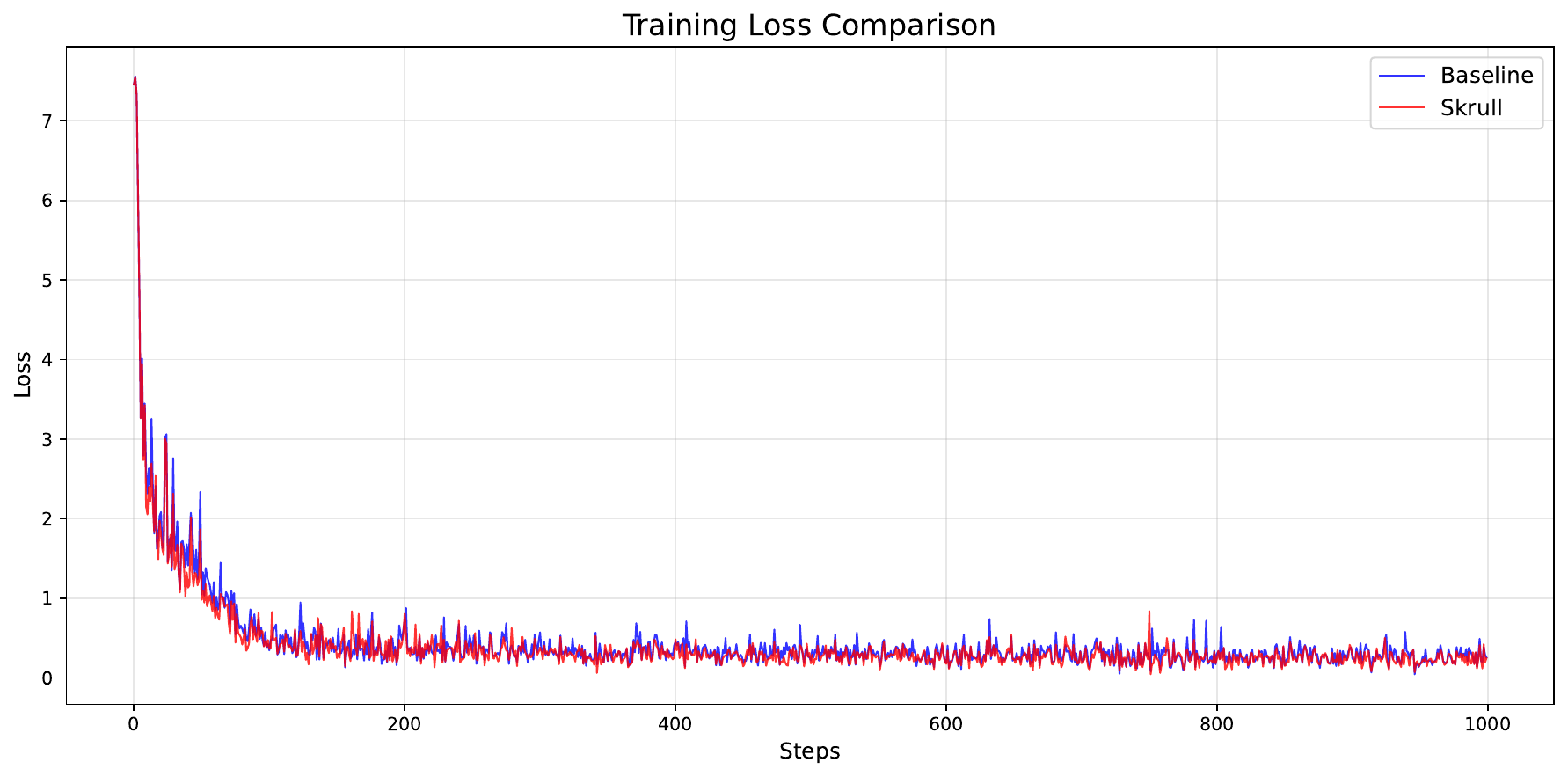}
    \caption{Loss comparison between Skrull and standard training method.}
    \label{fig:loss_curves}
\end{figure}

To evaluate the loss equivalence, we compare the loss carve between Skrull and standard training method when training Qwen2.5-0.5B on LMsysChat1M. The data scheduling in Skrull alter the accumulation order and we can observe slightly numerical differences due to the non-associativity of floating-point operations. However, Skrull do not alter any contents and orders in each global batch, the optimization trajectory remains equivalent. Therefore, as shown in Figure~\ref{fig:loss_curves}, Skrull does not influence the convergence.

\section{Heuristic Scheduling Algorithm}

\subsection{Function Definition of Heuristic Algorithm}
\label{apdx:details_on_heuristic_CP}
\begin{algorithm}[H]
\caption{Function Definations in scheduling algorithm for DACP}
\label{alg:heuristic_cp_detail}
\begin{algorithmic}[1]
\Require SeqNum $K$, SeqLens $S[K]$, Buckets $C$, CP degree $N$, Loads $L[N]$, RemainBucket $RB[N]$, DACP scheduling result $ret$

\Function{UpdateLocal}{$idx$, $rank$}
   \State $RB[rank] \gets RB[rank] - S[idx]$ \Comment{Update remaining bucket capacity}
   \State $L[rank] \gets L[rank] + FLOPs(S[idx])$   \Comment{Update current load}
\EndFunction

\Function{UpdateAll}{$idx$}
   \For{$i=0$ \textbf{to} $N-1$}
      \State $RB[i] \gets RB[i] - S[idx]/N$ \Comment{Distribute across all buckets}
      \State $L[i] \gets L[i] + FLOPs(S[idx],N)$   \Comment{Update all loads}
   \EndFor
\EndFunction

\Function{RollBack}{$rank$,$RB$,$L$}
    \For{$i=0$ \textbf{to} $K-1$}
        \If{$ ret[i] == rank $}
            \State $ ret[i] \gets -1$ \Comment{Distribute the sequence}
            \State $ RB[rank] \gets RB[rank] - S[i] + S[i]/N$
            \If{$L$ is not None}
                \State $ L[rank] \gets L[rank] - FLOPs(S[i]) + FLOPs(S[i],N)$
            \EndIf
            \State \Return $True$ \Comment{Success Roll-back}
        \EndIf
    \EndFor
    \State \Return $False$ \Comment{Roll-back Failed}
\EndFunction

\end{algorithmic}
\end{algorithm}

\subsection{Round-robin Scheduling Algorithm} 
\label{apdx:RR_scheduling}

\begin{algorithm}[h]
   \caption{Round-Robin Scheduling Algorithm}
   \label{alg:greedy_scheduling}
\begin{algorithmic}[1]  
   \State {\bfseries Input:} SeqNum $N$, SeqLens $S$, Buckets $C$, WorldSize $ws$
   \State {\bfseries Output:} PartitionIds $P$
   \For{$i=1$ {\bfseries to} $m-1$}
   \State $t \gets \text{FindMaxBucketsIds()}$
   \If{ $C[t] \geq S[i]$ } 
      \State $P[i] \gets t$ \hfill $\triangleleft$ \text{fit the max bucket} 
   \Else
      \State $j \gets \text{FindMinBucketsIds()}$
      \If{ $C[j] \geq S[i]/ws$}
         \State $P[i] \gets -1$ \hfill $\triangleleft$ \text{partition the sequence}
      \Else
         \State \textbf{Assert} RollBack$(j,C)$ \hfill \textcolor{black}{$\triangleleft$ \text{with roll-back}}
         \State $i \gets i - 1$
         \State \textbf{continue}
      \EndIf
   \EndIf
   \EndFor
\end{algorithmic}
\end{algorithm}


\section{Performance Modeling}
\label{sec:apdx_performance_modeling}
\subsection{Memory Estimation}
\label{sec:apdx_bucketsize}
Due to limited pages, we discuss the memory estimation methodology of Skrull in this section. The key point of this section is the determination of BucketSize $C$, which maps memory capacity to sequence token length. 

We first analyze the memory consumption during LLMs training. The memory consumption can be roughly categorized into two components: the static memory and the dynamic memory. The static memory, which typically includes model parameters and optimizer states, remains roughly constant throughout the training process given specific model configurations and parallelism strategies. In contrast, the dynamic memory or activation memory, varies with the input workload. In transformer architectures, activation memory is proportional to the sequence length. For instance, the Linear module, LayerNorm and \texttt{Attention} module (using FlashAttention \cite{flashattention_v1_2022, flashattention_v2_2024}) exhibit a linear relationship with sequence length. Therefore, we can estimate activation memory for a given sequence length $S$ using the following equation:
\begin{equation}
    Memory(S) = \alpha S + \beta 
\end{equation}
Here, the coefficient $\alpha$ and constant $\beta$ is determined at offline profiling. Notably, some memory reduction strategies, such as gradient checkpoints, only affect the $\alpha$ and $\beta$. We can still apply offline profiling method to estimate activation memory. In our implementation, we found that $\beta$ is usually negligible. Additionally, we employ sequence packing to eliminate padding and enhance performance, allowing us to directly use the total sequence length for memory estimation. Consequently, through offline profiling, we can deduce the BucketSize $C$ under various settings.

\subsection{Computation Estimation}
\label{sec:apdx_compuation_estimator}
In this section, we describe the methodology used to estimate the computational cost $T_{comp}$. 

Accurately modeling the computational cost as a function of sequence length $S$ is non-trivial. Simply assuming a linear or quadratic relationship with sequence length is insufficient because the computational FLOPs of TransformerLayer are dominated by the Linear and \texttt{Attention} modules, exhibiting a hybrid of linear and quadratic dependencies on $S$. The relative contributions of these components vary depending on the specific model configuration. Therefore, we formulate a function of FLOPs to provide roughly computational cost estimation given a specific model configuration and sequence length $S$.

Given the model configuration of hidden dimension $h$, key/value hidden dimension $h_{kv}$ and training batchsize $b$ (usually be 1 when employ sequence packing), the FLOPs is estimated as the Equation~\ref{eq:flops_func}. 

\begin{equation}
\label{eq:flops_func}
    FLOPs(S_k) = 20*b*h^2*S_k + 4*b*h*h_{kv}*S_k + 4*b*h*S_k^2
\end{equation}
For each sequence, the $T_{comp}$ can be estimated as:
\begin{equation}
\label{eq:T_cal}
    T_{comp} = \alpha FLOPs + \beta
\end{equation}
where all the $\alpha$ and $\beta$ is determined when offline profiling. 

\begin{figure}[H]
    \centering
    \includegraphics[width=0.5\linewidth]{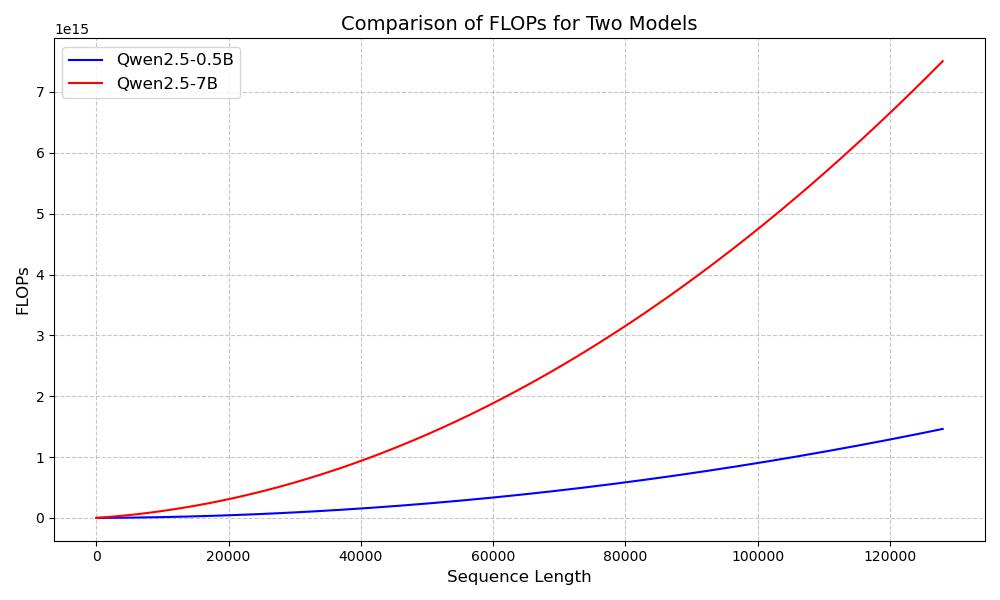}
    \caption{FLOPs VS Sequence Length on Qwen-2.5 0.5B and 7B}
    \label{apdx:fig:flops_length}
\end{figure}

Furthermore, as shown in Figure~\ref{apdx:fig:flops_length}, we plot the relationship between FLOPs and sequence length for Qwen-2.5-0.5B and Qwen-2.5-7B. The results highlight the distinct characteristics of long and short sequences. For short sequences, both computational workload and activation memory consumption scale roughly linearly with sequence length. However, for the long sequences, the computational workload grows rapidly due to the dominance of the quadratic term, while memory consumption still remains linear, leading to the problem of trade-off between balancing computation and memory, which is discussed in detail in Section~\ref{sec:efficient_solver} where we present insights into our heuristic algorithm design. 

Additionally, the transition point at which the quadratic term dominates varies depending on the model configuration. As demonstrated in Figure~\ref{apdx:fig:flops_length}, Qwen-2.5-7B, which has a larger hidden dimension $h$, exhibits a more rapid increase in FLOPs compared to Qwen-2.5-0.5B. Although Qwen-2.5-0.5B has slower FLOPs increase, we take it as example to further discuss the distinct characteristics between long and short sequences. In Qwen-2.5-0.5B, the quadratic term begins to dominate only when the sequence length $S$ exceeds approximately 4K, exhibits roughly linear relationship in short sequences. However, when $S=32K$, the total computational workload is 30 times greater than when $S=4K$, while the memory consumption increases only 4-fold. These estimations further elucidate the distinct characteristics of long and short sequences.

\subsection{Communication Estimation}
\label{sec:apdx_comm_estimator}
For the $T_{comm}$, we can simply profile in offline ways. Concretely, when the communication volume is smaller then a threshold, the fixed overhead of communication dominates the latency. However, with size increased, the fixed overhead become negligible and the latency is approximately proportional to communication volumes. We can deduce the thresholds, fixed overhead and the estimation function through a simple profiling. As shown in Table~\ref{tab:comm}, we plot the communication performance profiling results. Therefore, we can fit the Equation~\ref{eq:T_comm} according to communication volume $V$ in different hardware environments. Then, we can derive the communication volume according to sequence length $S$ under different model configurations as shown in Equation~\ref{eq:volume_fuc}, where $hidden_{kv}$ and $b$ means hidden dimension of Key/Value and batch size.
\begin{equation}
\label{eq:volume_fuc}
    Volume(S) = b*S_k*hidden_{kv}
\end{equation}

\begin{equation}
\label{eq:T_comm}
    T_{comm} = \left( \alpha V + T_{fixed} \right)
\end{equation}

\begin{table}[]
\caption{Collective Communication Latency Profiling.}
\label{tab:comm}
\centering
\begin{tabular}{c|c|c|c|c}
\hline
Size (MB)/Latency(us) & All\_gather & All\_to\_All & Reduce\_scatter & All\_reduce \\ \hline
2                     & 53.29       & 80.62        & 59.48           & 84.65       \\ \hline
4                     & 72.52       & 78.63        & 79.26           & 113.3       \\ \hline
8                     & 97.86       & 110.9        & 104.7           & 168.4       \\ \hline
16                    & 199.3       & 163.2        & 177.4           & 312.2       \\ \hline
32                    & 286.2       & 277.5        & 269.5           & 479.2       \\ \hline
64                    & 488.6       & 502.4        & 458.8           & 859.7       \\ \hline
128                   & 910.6       & 939.2        & 864.3           & 1642.9      \\ \hline
256                   & 1758.4      & 1803.9       & 1663.9          & 3197.9      \\ \hline
512                   & 3416.4      & 3411.2       & 3239.5          & 6181.2      \\ \hline
1024                  & 6467.9      & 6629.6       & 6294.3          & 12126       \\ \hline
\end{tabular}
\end{table}


\clearpage

\end{document}